\documentclass[preprint,12pt]{elsarticle}

\usepackage{amssymb}
\usepackage{amsmath}
\usepackage{mathrsfs}
\usepackage{dsfont}
\usepackage{booktabs}
\usepackage{subcaption}
\usepackage{hyperref}
\usepackage{cleveref}
\usepackage{color}
\usepackage{multirow}

\newcommand{\class}[1]{\mathcal{C}_{#1}}
\newcommand{\order}[1]{\mathcal{O}(#1)}

\journal{Computer Methods and Programs in Biomedicine}

\graphicspath{{img/}}

\begin{document}

\begin{frontmatter}



\title{From Kellgren–Lawrence to Calcium Pyrophosphate Crystal Deposition: A Soft-Labelling Framework for Knee Osteoarthritis Assessment}

\author[1]{Francisco Bérchez-Moreno}

\author[3]{Riccardo Rosati}
\author[5]{Maria Chiara Fiorentino}
\author[2]{Víctor M. Vargas \corref{cor1}}
\ead{victor.vargas@uah.es}
\author[6,7]{Edoardo Cipolletta}
\author[8]{Emilio Filippucci}
\author[4]{Luca Romeo}
\author[1]{Pedro A. Gutiérrez}
\author[1]{César Hervás-Martínez}

\cortext[cor1]{Corresponding author}

\affiliation[1]{organization={Departamento de Ciencia de la Computación e Inteligencia Artificial, Universidad de Córdoba},
            city={Córdoba},
            country={España}}

\affiliation[2]{organization={Departamento de Teoría de la Señal y Comunicaciones, Universidad de Alcala},
            city={Madrid},
            country={España}}
            
\affiliation[3]{organization={Department of Political Science, Communication, and International Relations, University of Macerata},
            city={Macerata},
            country={Italy}}
            
\affiliation[4]{organization={Department of Economics and Law, University of Macerata},
            city={Macerata},
            country={Italy}}

\affiliation[5]{organization={Department of Innovative Technologies in Medicine \& Dentistry, Università degli Studi "G. D'Annunzio" Chieti - Pescara},
            city={Chieti},
            country={Italy}}
            
\affiliation[6]{organization={Department of Internal Medicine, Azienda Ospedaliero Universitaria delle Marche}, city ={Ancona},  country={Italy}}

\affiliation[7]{organization={Academic Rheumatology, University of Nottingham}, city ={Nottingham}, country={UK}}

\affiliation[8]{organization={Department of Rheumatology, Polytechnic University of Marche}, city={Ancona}, country={Italy}}

\begin{highlights}
\item Curated X-ray dataset enables paired KL and CPPD severity analysis.
\item Ordinal soft-labelling models uncertainty between KL and CPPD grades.
\item Beta and triangular distributions significantly improve QWK and MAE.
\item Framework preserves ordinal structure and KL-CPPD scale correlations.
\end{highlights}

\begin{abstract}
\textbf{Background and objective.} Conventional Deep Learning (DL) approaches for Knee Osteoarthritis (KOA) grading rely on one-hot labels, which fail to capture both the ordinal uncertainty of Kellgren--Lawrence (KL) and Calcium Pyrophosphate Deposition Disease (CPPD) severity scores and the asymmetric relationship between the two scales observed in clinical practice.
\\
\textbf{Methods.} We retrospectively collected 2172 knee X-ray images, including 968 radiographs jointly annotated for KL and CPPD severity. An ordinal DL framework based on soft-labelling was developed for both tasks, replacing one-hot targets with unimodal probability distributions centred on the annotated grade. Four formulations were investigated: binomial, beta, triangular, and exponential.
\\
\textbf{Results.} All soft-labelling strategies consistently outperformed the nominal baseline. For CPPD grading, the triangular formulation achieved the highest Quadratic Weighted Kappa (QWK) and the lowest Mean Absolute Error (MAE) (QWK = $0.796_{0.018}$; MAE = $0.438_{0.025}$), while the beta formulation yielded the most balanced class-wise performance considering Average MAE (AMAE) and Maximum MAE (MMAE) across classes (AMAE = $0.458_{0.032}$; MMAE = $0.573_{0.084}$). For KL grading, the beta-based approach provided the best overall performance, achieving the highest QWK together with the lowest MAE and class-wise errors (QWK = $0.777_{0.022}$; MAE = $0.529_{0.031}$; AMAE = $0.523_{0.031}$; MMAE = $0.775_{0.136}$). Statistical analysis demonstrated significant improvements over conventional one-hot supervision ($p < 0.001$).
\\
\textbf{Conclusions.} Overall, the proposed ordinal soft-labelling framework more effectively captured the uncertainty and asymmetric relationship between KL and CPPD severity scales, supporting more reliable and clinically meaningful knee radiograph assessment.

\end{abstract}




\end{frontmatter}

\section{Introduction}\label{sec:introduction}

Osteoarthritis (OA) is the most prevalent degenerative joint disease and a leading cause of disability worldwide, particularly among the elderly population \cite{allen2022epidemiology}. Knee Osteoarthritis (KOA) represents the most common and clinically impactful manifestation, substantially impairing mobility and quality of life. Although not life-threatening, OA imposes a significant socioeconomic burden due to chronic pain, reduced functionality, and the progressive need for surgical interventions\cite{sharma2021osteoarthritis}.

Conventional radiography remains the primary diagnostic tool for assessing structural joint changes in OA \cite{sakellariou2017eular}. To standardise disease assessment, the Kellgren–Lawrence (KL) grading system is one of the most widely adopted semiquantitative scale for evaluating OA severity \cite{kohn2016classifications}. The KL score considers the presence and extent of osteophyte formation, joint space narrowing, subchondral sclerosis, and bone deformity, providing a structured framework for staging structural degeneration. 

However, degenerative joint changes do not occur in isolation. Crystal arthritis such as Calcium Pyrophosphate Crystal Deposition (CPPD) disease, may coexist with osteoarthritic alterations. CPPD disease is characterised by the accumulation of calcium pyrophosphate crystals in cartilage and periarticular tissues, frequently manifesting radiographically as chondrocalcinosis \cite{filippou2021critical}. Comprehensive knee radiographic evaluation should account for both structural degenerative changes and crystal deposition patterns to make a diagnosis (KOA, CPPD, KOA with CPPD). Importantly, periarticular calcifications may overlap or obscure classical OA features, potentially complicating KL grading. In addition, chondrocalcinosis is often overlooked and not reported in radiological reports. 

\begin{figure}[!ht]
    \centering
        \includegraphics[width=1\textwidth]{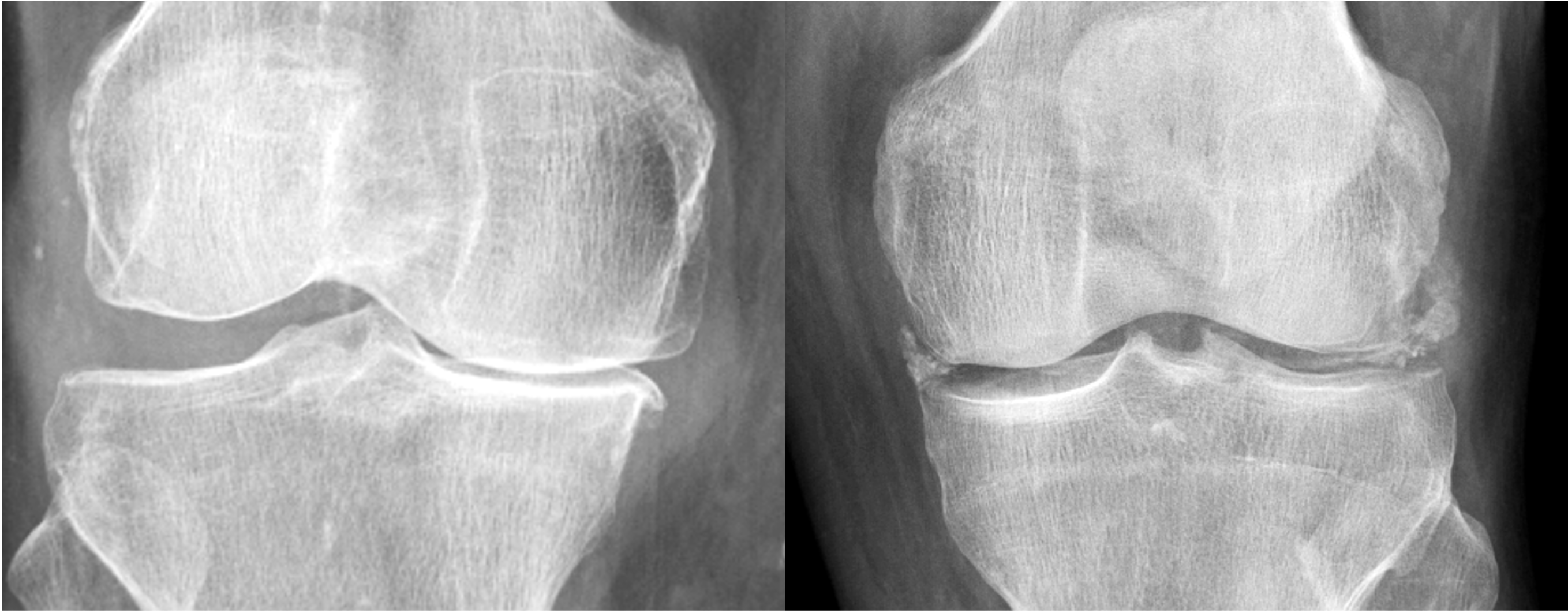}
    \caption{Knee radiographs illustrating the complexity of jointly assessing Osteoarthritis (OA) severity and Calcium Pyrophosphate Crystal Deposition (CPPD). Left: knee with KL grade 4 and CPPD score 0, showing severe joint space narrowing and prominent osteophyte formation, which may limit the visibility of potential calcifications. Right: knee with KL grade 2 and CPPD score 3, where the joint space is relatively preserved but largely occupied by calcific deposits consistent with chondrocalcinosis.}
    \label{fig:cppd}
\end{figure}

In this context, automated and reproducible algorithms capable of estimating KL grade and chondrocalcinosis extent are needed to reduce interpretative variability, improve consistency, and support routine clinical decision- making. Although KL grading is well standardised, CPPD scoring is less systematically defined. In addition, evaluating CPPD might be difficult in patients with severe OA (see Figure \ref{fig:cppd}).

Artificial Intelligence (AI) methodologies have been widely applied to KOA over the past decade \cite{buchlak2022clinical, teoh2023stratifying}. 
Significant recent studies include bone segmentation \cite{ahmed2022comprehensive}, automatic feature extraction via Convolutional Neural Networks (CNNs) \cite{si2022deep,lv2025lmsst}, and models trained on specific KOA datasets \cite{hinterwimmer2022machine}.
Despite this progress, current approaches predominantly rely on categorical cross-entropy or regression-based formulations. These paradigms treat severity grades as either unrelated categories or continuous values, disregarding two fundamental properties of radiographic scoring: the ordinal nature of the grading scales, in which misclassification severity grows with the distance between predicted and true grades, and the substantial inter-observer variability, which makes hard boundary assignments between adjacent grades inherently uncertain. 
These observations suggest that representing each ground-truth label as a deterministic one-hot vector is a poor modelling choice: a soft-labelling strategy based on unimodal probability distributions centred on the annotated grade can simultaneously encode ordinal proximity and label uncertainty, yielding supervision signals that are both more informative and more clinically faithful. Although recent work has introduced ordinal constraints to penalise large-rank errors \cite{chen2019fully, yong2022knee}, no study has combined unimodal soft labels with regularised ordinal loss functions within a unified framework.
Furthermore, the integration of multi-score assessments, specifically KL and CPPD grading, remains largely unexplored. Although these grading systems often coexist in clinical practice, they capture complementary pathological aspects and may not progress in a fully consistent or linear manner. This makes it inappropriate both to model them independently and to assume a direct one-to-one correspondence between them. Consequently, there is a need for computational approaches able to model KL and CPPD severity while preserving the ordinal structure of each scale and accounting for the relationships observed between them.

The main contributions of this work can be summarised as follows:
\begin{itemize}
    \item We acquired and curated a dedicated dataset comprising paired KL and CPPD assessments, enabling the study of their co-occurrence and relationship in a clinically realistic setting.

    \item We introduce an ordinal soft-labelling framework based on unimodal probability distributions and regularised loss functions, explicitly designed to model the gradual transition between neighbouring severity grades while penalising large-rank errors more strongly. Unlike conventional one-hot supervision, the proposed formulation captures the uncertainty and overlap inherent to adjacent KL and CPPD categories. A comprehensive experimental analysis is conducted to compare multiple soft-labelling distributions, loss functions, and regularisation strategies across both grading tasks, thereby identifying the most suitable supervision scheme for KOA severity assessment.

    \item We show that, although KL and CPPD are predicted through separate models, the proposed soft-labelling formulation preserves the ordinal structure of both grading systems and encourages predictions that remain coherent with the correlations between the two scales observed in the training data. In this way, the framework captures their complementary yet related pathological progression, without imposing an unrealistic direct correspondence between KL and CPPD grades.

\end{itemize}

The remainder of the paper is organised as follows. \Cref{sec:related-work} reviews related work. \Cref{sec:proposed-methodology} presents the dataset, the proposed methodology, experimental design and model configuration. \Cref{sec:results} reports the experimental results. \Cref{sec:discussion} discusses the experimental results from an expert point of view. Finally, \Cref{sec:conclusions} concludes the paper.

\section{Related work}\label{sec:related-work}

The automated assessment of KOA severity has undergone substantial progress with the advent of DL techniques \cite{kokkotis2020machine}. CNNs have become the dominant paradigm for automatic bone segmentation \cite{ahmed2022comprehensive} and feature extraction from radiographic images \cite{si2022deep}, enabling increasingly accurate prediction of disease severity according to the KL grading system. Most state-of-the-art methods rely on CNN-based architectures to classify KOA severity into discrete KL grades \cite{hinterwimmer2022machine, upadhyay2023detection}. Some studies further improve performance through dedicated preprocessing strategies, including frequency-domain filtering to better capture trabecular bone texture \cite{wang2022resnet}, or introducing weighted loss functions and metrics to reflect the greater clinical relevance of severe misclassifications \cite{brejnebol2022external, chaugule2022knee}. 

For instance, \cite{kalpana2023evaluating} systematically compared 12 pre-trained CNN architectures, including VGG, ResNet, DenseNet, EfficientNet, and MobileNet, identifying MobileNet as the best-performing model. More recently, transformer-based approaches have also been explored to better capture both local and global image characteristics. In particular, \cite{jahan2024koa} proposed KOA-CCTNet, an enhanced framework based on a modified Compact Convolutional Transformer (CCT) that combines convolutional layers with transformer modules to improve feature representation and KOA grade prediction. Likewise, \cite{maqsood2025knee} introduced a hybrid transformer-based architecture integrating CNN-based local feature extraction with transformer-based global context modelling for KOA detection and grading.

\subsection{Soft-labelling}\label{subsec:soft-labelling}

In the medical domain, Ordinal Classification (OC) is particularly relevant whenever disease severity is described through discrete but inherently ordered clinical stages. Representative applications include cancer grading \cite{albuquerque2021ordinal, le2021joint}, Alzheimer's disease progression \cite{wang2024novel}, donor--recipient allocation in liver transplantation \cite{rivera2023ordinal}, and the assessment of cerebral collateral status in acute ischemic stroke \cite{le20223d}. In medical imaging, explicitly incorporating ordinal information during training has been shown to reduce large misclassification errors between distant categories, which are generally less acceptable from a clinical perspective \cite{chen2019fully, yong2022knee}.

A common strategy to incorporate ordinal relationships into DL models is soft-labelling, where the hard one-hot target is replaced by a probability distribution that assigns higher confidence to neighbouring classes and lower confidence to distant ones. In this way, the model is encouraged to learn the gradual transition between adjacent severity levels rather than treating all classes as equally distinct, thereby accounting for both labelling uncertainty and the similarity between adjacent classes.

Several approaches have been proposed to generate these soft targets. Early works focused on imposing unimodal constraints on the predicted label distribution. For example, \cite{liu2020unimodal} introduced a unimodal-regularised stick-breaking process, yielding smooth and consistent ordinal predictions. Similarly, \cite{li2022unimodal} proposed the unimodal-concentrated loss, which dynamically generates sample-specific label distributions during training to improve robustness to ambiguity and noisy labels. In the same study, the authors compared different probability distributions, including Poisson, binomial, and exponential distributions, for encoding ordinal labels.

More recently, different continuous distributions have been explored to construct soft labels with specific properties. In \cite{vargas2022unimodal}, the beta distribution was employed to generate smooth and unimodal label distributions for image-based ordinal tasks. This idea was subsequently extended through the use of exponential distributions \cite{vargas2023exponential}, which are particularly suitable when extreme classes are rare but clinically important, and triangular distributions \cite{vargas2023soft}, which provide a simple yet effective way to preserve ordinal relationships. To further increase flexibility, \cite{vargas2023generalised} proposed a generalised triangular distribution capable of adapting more closely to the underlying data distribution. A recent review by \cite{cardoso2025unimodal} provides a comprehensive overview of these unimodal soft-labelling strategies, discussing both their theoretical foundations and practical implications for ordinal classification. Another recent experimental review \cite{vargas2026soft} systematically analysed the impact of different loss functions, output layers, and labelling strategies in deep ordinal classification, showing that soft-labelling approaches generally improve generalisation performance across diverse ordinal imaging datasets.

Despite these advances, soft-labelling approaches have not yet been investigated in the context of knee osteoarthritis grading. Existing KOA studies still rely on hard labels, both for the traditional KL grading system and for alternative clinical scores. In particular, no previous work has explored soft-labelling strategies for the CPPD score, despite its ordinal nature and its relevance for characterising crystal-related arthropathy.

Therefore, in this work, we investigate soft-labelling for both KL and CPPD severity assessment, with particular emphasis on CPPD, for which no previous DL studies are available. Unlike prior methods that adopt a single distribution for all classes, we introduce a hybrid fusion strategy in which different probability distributions are assigned to extreme and intermediate classes. This formulation offers greater flexibility in modelling the progression between severity stages and provides a representation that more closely reflects the underlying clinical evolution of KOA and CPPD-related disease.

\section{Methodology}\label{sec:proposed-methodology}
This section begins by describing the curated KOA dataset employed in this study, including both KL and CPPD annotations (Section \ref{subsec:data-collection}). The joint analysis of these annotations provides insight into the relationship between the two grading systems and motivates the need for a soft-labelling ordinal formulation. Next, the proposed soft-labelling strategy and the associated ordinal learning framework are presented (\Cref{soft_labelling}). Finally, the training protocol, hyperparameter optimisation procedure, and evaluation metrics are described in \Cref{subsec:models_training,subsec:performance-metrics}, respectively.

\begin{figure}[!ht]
    \centering
        \includegraphics[width=1\textwidth]{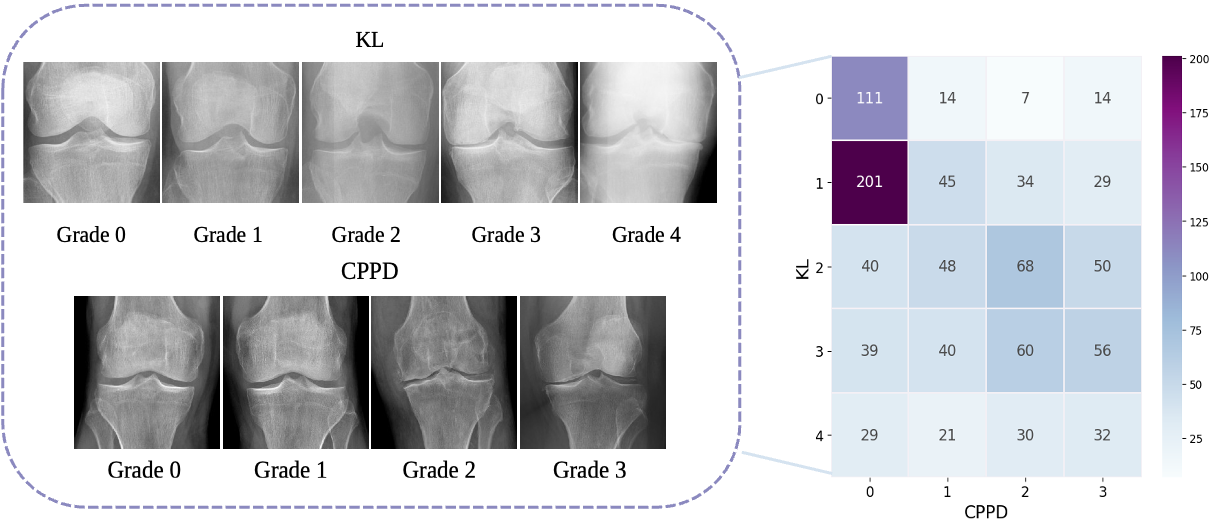}
    \caption{Representative examples of the different Kellgren--Lawrence (KL) and Calcium Pyrophosphate Deposition Disease (CPPD) severity grades and their relationship in the study cohort. On the left, example radiographs illustrating KL grades 0--4 (top row) and CPPD grades 0--3 (bottom row). On the right, contingency table showing the joint distribution of KL and CPPD grades in the dataset, where each cell reports the number of radiographs assigned to the corresponding combination of KL and CPPD severity.}
    \label{fig:ground_truth}
\end{figure}

\subsection{Data collection and processing}\label{subsec:data-collection}

Standard anteroposterior knee radiographs acquired in both supine and standing positions were retrospectively collected from the Radiology Departments of Carlo Urbani Hospital, Jesi (January 2023--June 2025), and Azienda Ospedaliero Universitaria delle Marche (August 2023--March 2025). Exclusion criteria included patient age below 18 years, previous knee arthroplasty, the presence of fixation devices, and radiographic evidence of fractures involving the knee or adjacent bones.

After applying these criteria, a total of 2,172 knee radiographs was retained. For most subjects, both left and right knees were available, resulting in a heterogeneous cohort characterised by substantial variability in acquisition conditions, patient positioning, radiographic quality, and disease severity.

Among the collected radiographs, 968 images were jointly annotated according to both the KL and CPPD grading systems, whereas the remaining 1,204 images were annotated exclusively for CPPD severity. The KL score was assigned on a five-level ordinal scale ranging from grade 0 (absence of radiographic signs of osteoarthritis) to grade 4 (severe osteoarthritis with marked joint space narrowing and bone deformity). CPPD severity was instead evaluated on a four-level scale ranging from grade 0 (absence of calcific deposits) to grade 3 (extensive and diffuse crystal deposition).

The availability of paired KL and CPPD annotations enabled the investigation of the relationship between structural osteoarthritic degeneration and crystal deposition patterns. As shown in \Cref{fig:ground_truth}, CPPD deposits are not confined to advanced osteoarthritis cases, but are also present in knees with mild or even absent radiographic osteoarthritic changes. Conversely, severe KL grades are not always associated with high CPPD severity. Moreover, the dataset presents several additional challenges that complicate automatic grading. First, the number of samples is unevenly distributed across severity grades, with KL grade~1 and CPPD grade~0 being the most represented classes, whereas severe KL grade~4 is substantially less frequent. Second, the radiographic appearance of adjacent grades often overlaps considerably: for example, the distinction between KL grades~1 and~2 or between CPPD grades~1 and~2 may be subtle and depend on minimal changes in joint space narrowing, osteophyte formation, or calcification visibility. Finally, the coexistence of osteoarthritic changes and CPPD deposits may generate confounding visual patterns, as extensive osteophytes, sclerosis, and joint narrowing can partially obscure or mimic CPPD-related calcifications.

To illustrate these challenges, \Cref{fig:ground_truth} reports representative examples of each KL and CPPD grade together with the contingency table derived from the subset of jointly annotated images. The distribution reveals a clear asymmetric association between the two grading systems. At lower severity levels, there is a strong correspondence: for KL $0$ and $1$, the counts are heavily concentrated in the CPPD $0$ category ($111$ and $201$ samples, respectively). However, this alignment disappears as KL severity increases. For the highest grade, KL $4$, the cases are distributed almost uniformly across all CPPD categories (ranging from $21$ to $32$), showing that a high KL score does not necessarily imply a high CPPD score. Furthermore, this lack of symmetry is evident in the extreme discordant grades: the combination of high KL with low CPPD ($29$ cases) is twice as frequent as low KL with high CPPD ($14$ cases). These observations confirm that the progression of CPPD and KL does not follow a symmetric one-to-one relationship.
 
To reduce irrelevant anatomical context and focus the analysis on the knee joint region, all radiographs underwent a preprocessing step based on automatic localisation. Specifically, a YOLOv8 model was used to identify the knee region and remove superfluous areas. To train this detector, a subset of 100 images was manually annotated by drawing bounding boxes around the knee joint. The trained model was then used to automatically generate bounding boxes for the remaining images. In this way, all images were cropped to the knee region. Given that the size of the bounding boxes produced by the YOLOv8 model can vary, the resulting images were resized to a fixed spatial resolution of $224 \times 224$ pixels prior to model training.

The class distributions for the KL and CPPD grading systems in the training and test sets are summarised in \Cref{tab:dataset_kl,tab:dataset_cppd}. 

\begin{table}[!t]
\centering
\scriptsize
\caption{Distribution of Kellgren--Lawrence (KL) severity grades in the training and test sets.}
\label{tab:dataset_kl}
\renewcommand{\arraystretch}{1.15}
\setlength{\tabcolsep}{10pt}
\begin{tabular}{p{8.5cm}cc}
\toprule\toprule
\textbf{KL severity grade} & \textbf{Training set} & \textbf{Test set} \\
\midrule
0 -- No radiographic signs of osteoarthritis 
    & 102 & 44 \\
1 -- Doubtful narrowing of the joint space and possible osteophytic lipping 
    & 217 & 93 \\
2 -- Mild osteoarthritis with definite osteophytes and possible joint space narrowing 
    & 145 & 62 \\
3 -- Moderate osteoarthritis with multiple osteophytes, definite joint space narrowing, and possible bony deformity 
    & 137 & 58 \\
4 -- Severe osteoarthritis with marked joint space narrowing, large osteophytes, severe sclerosis, and definite bony deformity 
    & 78 & 34 \\
\midrule
\textbf{Total} & \textbf{679} & \textbf{291} \\
\bottomrule\bottomrule
\end{tabular}
\end{table}

\begin{table}[!t]
\centering
\scriptsize
\caption{Distribution of Calcium Pyrophosphate Deposition Disease (CPPD) severity grades in the training and test sets.}
\label{tab:dataset_cppd}
\renewcommand{\arraystretch}{1.15}
\setlength{\tabcolsep}{10pt}
\begin{tabular}{p{8.5cm}cc}
\toprule\toprule
\textbf{CPPD severity grade} & \textbf{Training set} & \textbf{Test set} \\
\midrule
Grade 0 -- No visible CPPD deposits 
    & 571 & 245 \\
Grade 1 -- Mild CPPD deposition with small or sparse calcifications 
    & 260 & 112 \\
Grade 2 -- Moderate CPPD deposition with evident calcifications involving a larger joint area 
    & 336 & 144 \\
Grade 3 -- Severe CPPD deposition with extensive and clearly visible calcifications 
    & 352 & 150 \\
\midrule
\textbf{Total} & \textbf{1519} & \textbf{651} \\
\bottomrule\bottomrule
\end{tabular}
\end{table}

\subsection{Proposed framework}\label{soft_labelling}

\begin{figure}[!ht]
    \centering
    \includegraphics[width=1\textwidth]{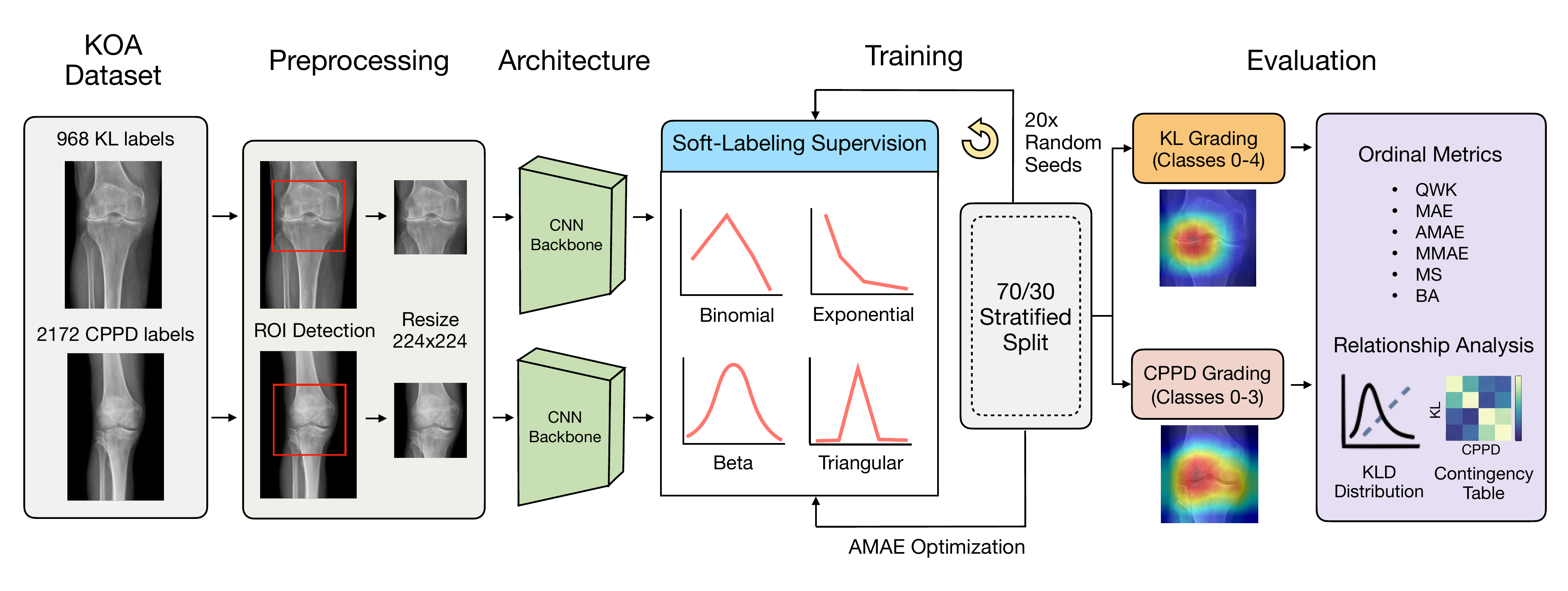}
    \caption{Schematic representation of the proposed ordinal deep learning framework for knee osteoarthritis (KOA) assessment . The pipeline includes automated knee localization via YOLOv8 and image resizing. A ResNet18 backbone is supervised using a soft-labelling framework based on four unimodal distributions (Binomial, Beta, Triangular, and Exponential) to capture ordinal uncertainty. Training is conducted using a 70/30 stratified split optimized for AMAE across 20 independent seeds. Final performance for KL (grades 0–4) and CPPD (grades 0–3) is evaluated using ordinal metrics and KLD-based relationship analysis.}
    \label{fig:framework}
\end{figure}

The proposed framework, depicted in \Cref{fig:framework}, is designed to explicitly account for the ordinal nature of KOA severity assessment. In contrast to most existing approaches, the proposed methodology does not alter the underlying CNN architecture, but instead modifies the supervision strategy used during training. This choice is motivated not only by the ordinal nature of the problem, but also by the increasing availability of powerful general-purpose visual backbones \cite{van2025foundation}. As modern CNN and transformer architectures already provide highly effective feature representations, further architectural modifications often yield only marginal improvements \cite{elharrouss2025vits}. Consequently, acting at the supervision level represents a more flexible and general strategy, allowing the proposed formulation to be integrated with different backbone architectures while remaining focused on the clinically relevant modelling of disease severity.

Formally, ordinal classification problems \cite{gutierrez2015ordinal} can be described as those in which the goal is to categorise patterns according to a discrete scale reflecting a natural order among labels. Let $\mathcal{Y} = \{\class{1}, \class{2}, \ldots, \class{J}\}$ denote the set of $J$ ordered categories arranged according to the relation $\class{1} \prec \class{2} \prec \cdots \prec \class{J}$, where $\prec$ represents the intrinsic ordering defined by the application domain.

Given an input space $\mathcal{X} \subseteq \mathds{R}^d$, where $d$ denotes the dimensionality of the feature space, each observation is represented by an input $\mathbf{x}_i \in \mathcal{X}$ and associated with a label $y_i \in \mathcal{Y}$. In image-based applications, $d$ is typically defined as $H \times W \times C$, where $H$ and $W$ denote the image height and width, respectively, and $C$ the number of channels. In this setting, $\mathbf{x}_i$ corresponds to a three-dimensional tensor rather than a vector; however, for notational simplicity, we retain the same formulation. A supervised ordinal classification dataset is therefore represented as $\mathcal{D} = \langle \mathbf{X}, \mathbf{y} \rangle = \{(\mathbf{x}_i, y_i)\}_{i=1}^{N}$, where $\mathbf{X} = \langle \mathbf{x}_1, \ldots, \mathbf{x}_N \rangle$ denotes the collection of input samples, $\mathbf{y} = \langle y_1, \ldots, y_N \rangle$ the corresponding labels, and $N$ the total number of samples. The learning task consists of constructing a predictive model that, given a new instance, outputs an estimated label $\hat{y}_i \in \mathcal{Y}$. Different encodings can be used to represent class labels, among which one-hot encoding is the most common. Under this formulation, the categorical cross-entropy (CCE) loss is defined as:

\begin{equation}\label{eq1}
    \mathscr{L}(\mathbf{x}_i, k) = \sum^J_{j=1} p(j, k)\left(-\log \, \hat{p}_j(\mathbf{x}_i)\right),
\end{equation}
where $k$ denotes the index of the ground-truth label of the $i$-th sample, and $\hat{p}_j(\mathbf{x}_i)$ is the probability predicted by the model that instance $\mathbf{x}_i$ belongs to class $\class{j}$. In the one-hot setting, the target distribution is given by $p(j,k) = \mathds{1}(y_i=\class{k})$, being $\mathds{1}(\cdot)$ the indicator function, such that $\mathds{1}(y_i=\class{k})=1$ if the label of the $i$-th is $\class{k}$, and $\mathds{1}(y_i=\class{k})=0$ otherwise. Therefore, all the target probability mass is concentrated on the annotated class, and all misclassifications are penalised equally regardless of their distance from the true grade.
However, this assumption is poorly suited to both KL and CPPD scoring, where the distinction between neighbouring grades is often subtle and affected by inter-observer variability \cite{moon2025deep}.

In the case of nominal soft-labelling, the CCE loss function is expressed as:
\begin{equation}\label{eq2}
    \mathscr{L}(\mathbf{x}_i,k) = \sum^J_{j=1} h_{SLN}(j,k) \left(-\log \, \hat{p}_j(\mathbf{x}_i)\right)
\end{equation}
where 
\begin{equation}
    h_\text{SLN}(j,k) = (1-\lambda)\mathds{1}(y_i=\class{k}) + \lambda \frac{1}{J}, \quad \lambda \in [0,1]
\end{equation}
represents a linear combination between the one-hot label and a uniform distribution, where $\lambda$ is a smoothing parameter that must be tuned and regulates how much probability mass is spread to other classes,

This combination of hard- and soft-labelling introduces a regularisation term in the loss function for a nominal classifier. In the case of ordinal classification, the uniform distribution in the second term of the summation is replaced by unimodal distributions, as it is assumed that misclassification errors are more likely to occur in classes adjacent to the true class. Thus, $h_{SLN}(j,k)$ in \ref{eq2} is replaced by:
\begin{equation}\label{eq3}
    h_\text{SLO}(j,k) = (1-\lambda)\mathds{1}(j=k) + \lambda \, \text{P}\left(\text{y} = \class{j} | \class{k}\right), \quad \lambda \in [0,1]
\end{equation}
where $\text{P}\left(\text{y} = \class{j} | \class{k}\right)$ is usually computed either from the probability mass function (p.m.f.) of a discrete distribution or from the probability density function (p.d.f.) a continuous one. When a continuous distribution is used, this probability can be written as:
\begin{equation}\label{eq4}
    \text{P}\left(\text{y} = \class{j} | \class{k}\right) = \int^{j/J}_{(j-1)/J} f(z,k)dz
\end{equation}
where $f(z,k)$ is the p.d.f. of a continuous probability distribution, associated with class $\class{k}$. The integration bounds, $(j-1)/J$ and $j/J$, divide the unit interval $[0,1]$ into $J$ equal-length segments, each corresponding to one of the $J$ classes, as described in \cite{vargas2022unimodal}.

\begin{figure}[!ht]
    \centering
    \includegraphics[width=\textwidth]{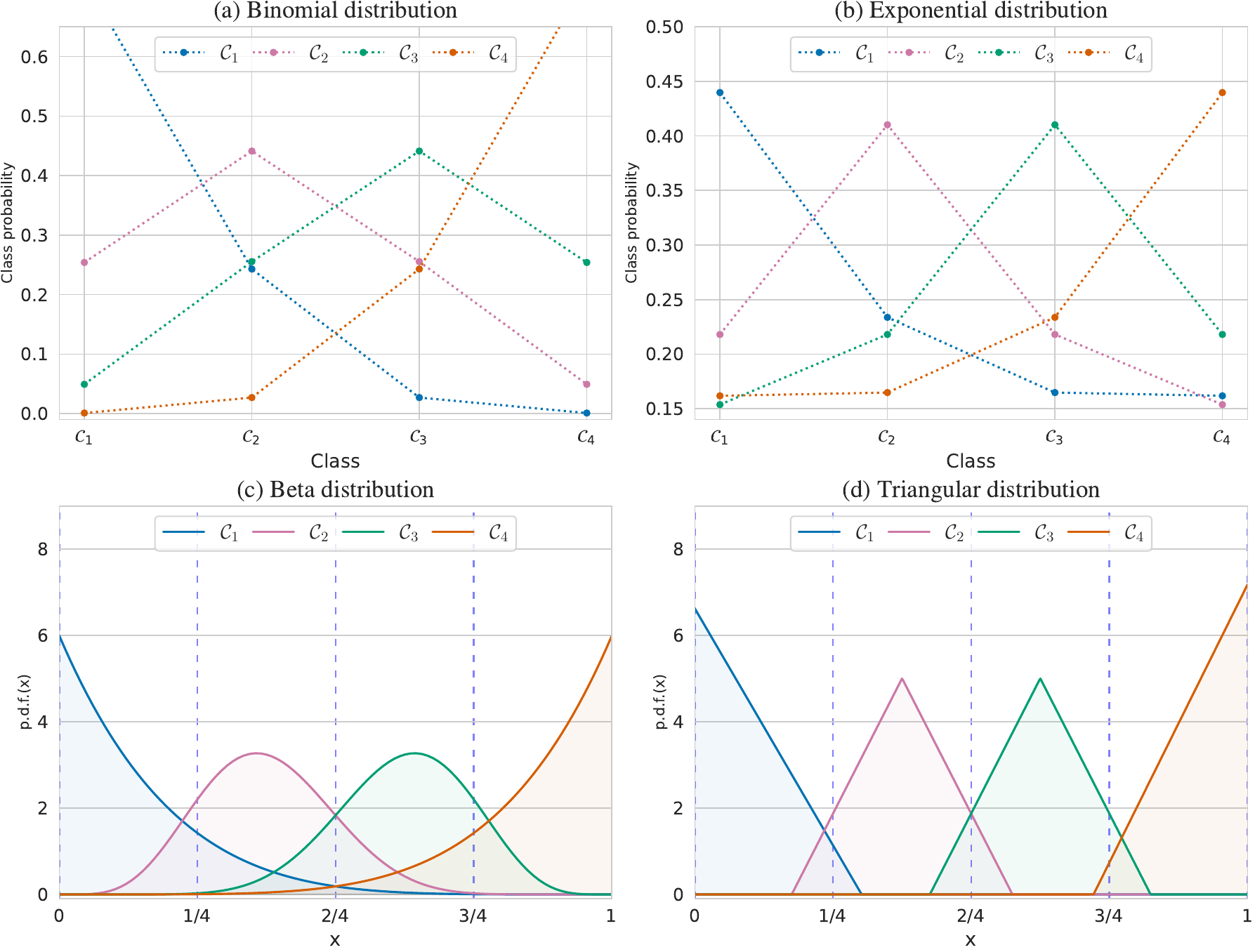}
    \caption{Example of the binomial (a), exponential (b), beta (c) and triangular (d) discrete and continuous unimodal distributions to obtain the soft labels of CPPD (4 classes or grades of severity)}
    \label{fig:unimodal_distributions}
\end{figure}

Within this study, four representative unimodal distributions are considered to define the conditional probability $\text{P}\left(\text{y} = \class{j} | \class{k}\right)$: triangular \cite{vargas2023soft}, beta \cite{vargas2022unimodal}, exponential \cite{vargas2023exponential}, and binomial formulations \cite{liu2020unimodal} (see \Cref{fig:unimodal_distributions}). These distributions were selected because they encode different assumptions about how uncertainty should be distributed around the annotated severity grade. In particular:
\begin{itemize}
    \item Triangular distribution \cite{vargas2023soft}: assigns the highest probability to the annotated class and decreases linearly towards neighbouring grades. This formulation naturally reflects the idea that confusion is more likely to occur between adjacent severity levels than between distant ones.

    \item Binomial distribution \cite{liu2020unimodal}: defines a discrete unimodal target centred on the ground-truth class, with probability progressively decreasing according to ordinal distance. Compared with the triangular formulation, it generally produces a more concentrated distribution around the annotated label.

    \item Beta distribution \cite{vargas2022unimodal}: introduces two shape parameters that regulate the concentration and spread of the probability mass. This additional flexibility makes it particularly suitable for modelling different degrees of uncertainty and overlap between adjacent grades.

    \item Exponential distribution \cite{vargas2023exponential}: models a distance-dependent decay controlled by a tunable $L_p$-norm parameter, allowing the transition between grades to be either sharper or smoother depending on the underlying severity progression.
\end{itemize}

The comparison between these distributions is particularly relevant in the present setting, as the degree of uncertainty is unlikely to be uniform across severity grades. Indeed, neighbouring categories often exhibit partially overlapping radiographic characteristics, whereas distant grades are generally easier to distinguish. Consequently, different soft-labelling formulations may better capture different patterns of disease progression and annotation ambiguity. 

\begin{table}[!t]
    \centering
    \caption{Hyperparameter search space considered for each learning strategy.}
    \label{tab:crossvalidation}
    \small
    \begin{tabular}{lll}
        \toprule\toprule
        \textbf{Method} & \textbf{Hyperparameter} & \textbf{Values} \\
        \midrule
        Nominal 
            & Learning rate 
            & $\{10^{-4},\,10^{-3},\,10^{-2}\}$ \\
        \midrule
        \multirow{2}{*}{Binomial, Beta} 
            & Learning rate 
            & $\{10^{-4},\,10^{-3},\,10^{-2}\}$ \\
            & Smoothing factor $\eta$ 
            & $\{0.8,\,1.0\}$ \\
        \midrule
        \multirow{3}{*}{Triangular} 
            & Learning rate 
            & $\{10^{-4},\,10^{-3},\,10^{-2}\}$ \\
            & Adjacent-class probability 
            & $\{0.01,\,0.05,\,0.10\}$ \\
            & Smoothing factor $\eta$ 
            & $\{0.8,\,1.0\}$ \\
        \midrule
        \multirow{3}{*}{Exponential} 
            & Learning rate 
            & $\{10^{-4},\,10^{-3},\,10^{-2}\}$ \\
            & Smoothing factor $\eta$ 
            & $\{0.8,\,1.0\}$ \\
            & Exponent $p$ 
            & $\{1.0,\,1.5,\,2.0\}$ \\
        \bottomrule\bottomrule
    \end{tabular}
\end{table}

\subsection{Models training}\label{subsec:models_training}
Each model configuration was trained separately for KL and CPPD prediction in order to account for the distinct ordinal structure of the two grading systems. To reduce the effect of stochastic variability and obtain robust performance estimates, all experiments were repeated over 20 independent random seeds. A ResNet18 architecture \cite{he2016deep}, initialised with ImageNet pretrained weights and implemented in PyTorch, was trained using a batch size of 128 for up to 100 epochs. Random seeds were used to control both the final-layer initialisation and dataset resampling.
 
A stratified random holdout split was first generated, reserving 70\% of the data for training and 30\% for testing. For each seed, the dataset was resampled while preserving the original stratification and train/test proportions. For every soft-labelling formulation, hyperparameters were optimised through a randomised search procedure, sampling a maximum of 15 configurations from the search space reported in \Cref{tab:crossvalidation}. Model selection was then performed on a validation set obtained from the training data (30\%), using the Average Mean Absolute Error (AMAE) as the optimisation criterion \cite{gomez2024orfeo}. AMAE is particularly suitable for ordinal and imbalanced problems because it computes the prediction error independently for each class and then averages the obtained values, preventing dominant classes from disproportionately influencing model selection:
\begin{equation}\label{eq:amae}
    \text{AMAE} = \frac{1}{J} \sum_{j=1}^J \frac{1}{N_j} \sum_{i=1}^{N_j} \left| \order{y_i} - \order{\hat{y}_i} \right|,
\end{equation}
where $J$ denotes the number of ordinal categories, $N_j$ is the number of samples belonging to class $\class{j}$, $\order{y_i}$ and $\order{\hat{y}_i}$ indicate the true and predicted ordinal positions of the $i$-th sample, respectively, (i.e. $\order{\class{k}}=k$). Lower values of AMAE correspond to more accurate predictions and fewer large-rank errors.

The validation set used for hyperparameter search was also employed to monitor the training process and apply an early stopping strategy when the validation loss did not decrease for 40 consecutive epochs.

All experiments were implemented using the \texttt{dlordinal} Python library \cite{berchez2025dlordinal}, which provides a unified framework for deep ordinal classification and is publicly available on GitHub\footnote{\url{https://github.com/ayrna/dlordinal}}.

\subsection{Evaluation metrics}\label{subsec:performance-metrics}

\subsubsection{Model evaluation}\label{subsec:performance-model}
To comprehensively evaluate the proposed ordinal classifiers, six complementary performance metrics are considered. Since the problem is characterised by both ordinal structure and class imbalance, the selected metrics assess not only the overall classification accuracy, but also the magnitude and distribution of misclassification errors across severity grades.

The Quadratic Weighted Kappa (QWK) \cite{de2018weighted, cohen1968weighted, warrens2012cohen} is adopted as the primary agreement measure, as it explicitly penalises errors according to their ordinal distance. It is defined as:
\begin{equation}\label{eq7}
    \text{QWK} = 1 - \frac{\sum\limits_{i=1}^{J} \sum\limits_{j=1}^{J} \omega_{ij} O_{ij}}{\sum\limits_{i=1}^{J} \sum\limits_{j=1}^{J} \omega_{ij} E_{ij}},
\end{equation}
where $J$ denotes the number of ordinal categories, $\mathbf{O}$ is the observed confusion matrix, and $\mathbf{E}$ is the expected confusion matrix under random assignment. The entries $\omega_{ij}$ quantify the penalty associated with predicting class $\class{j}$ when the true class is $\class{i}$, and are defined as:
\begin{equation}\label{eq8}
    \omega_{ij} = \frac{|i - j|^n}{(J - 1)^n},
\end{equation}
where $n$ controls the severity of the penalty. In this work, the commonly adopted quadratic formulation ($n=2$) is used, thereby assigning substantially larger penalties to distant misclassifications. The elements of the expected confusion matrix are computed as:
\begin{equation}\label{eq9}
    E_{ij} = \frac{O_{i \bullet} O_{\bullet j}}{N},
\end{equation}
where $O_{i \bullet}$ and $O_{\bullet j}$ correspond to the marginal sums of the observed confusion matrix.

The Mean Absolute Error (MAE) \cite{willmott2005advantages, baccianella2009evaluation} is also employed to directly quantify the average ordinal distance between the predicted and true grades:
\begin{equation}\label{eq10}
    \text{MAE} = \frac{1}{N} \sum_{i,j=1}^{J} |i-j|\, O_{ij},
\end{equation}
where $N$ is the total number of samples, $J$ is the number of ordinal categories, and $O_{ij}$ is the number of instances belonging to class $i$ that are predicted as class $j$. Unlike standard accuracy, MAE takes into account the magnitude of the error, assigning larger penalties to predictions farther from the correct severity level.

Because the dataset is imbalanced, AMAE is additionally considered (Eq.~\ref{eq:amae}). To further quantify the worst-case behaviour of the classifier, the Maximum MAE (MMAE) \cite{cruz2014metrics} is used:
\begin{equation}\label{eq12}
    \text{MMAE} = \max \left\{ \frac{1}{N_j} \sum_{i=1}^{N_j} \left| \order{y_i} - \order{\hat{y}_i} \right|; \, j = 1,\ldots,J \right\},
\end{equation}
where $N_j$ denotes the number of samples belonging to class $\class{j}$. MMAE therefore captures the largest class-specific error and is particularly useful for identifying whether a model performs poorly on minority or clinically relevant severity grades.

To complement these ordinal metrics, Minimum Sensitivity (MS) \cite{FernandezCaballero2010} is adopted to assess the worst-case sensitivity across classes:
\begin{equation}\label{eq13}
    \text{MS} = \min \left\{ \frac{O_{jj}}{O_{j\bullet}}; \, j = 1,\ldots,J \right\},
\end{equation}
where $O_{jj}$ denotes the number of correctly classified instances of class $j$, and $O_{j\bullet}$ is the total number of samples belonging to that class. MS is particularly relevant in medical applications, since it ensures that acceptable performance is achieved even for the least accurately predicted category.

Finally, Balanced Accuracy (BA) \cite{vargas2024ebano} is included to evaluate the overall classification performance while compensating for class imbalance:
\begin{equation}\label{eq14}
    \text{BA} = \frac{1}{J} \sum_{j=1}^{J} \frac{n_{jj}}{n_j},
\end{equation}
where $n_{jj}$ is the number of correctly classified samples in class $j$, and $n_j$ is the total number of samples belonging to that class. Unlike standard accuracy, BA gives equal importance to all severity grades, independently of their frequency in the dataset.

To qualitatively evaluate the models internal representations, we leverage Gradient-weighted Class Activation Mapping (Grad-CAM) \cite{selvaraju2017grad}. This visual interpretability analysis serves two fundamental purposes: i) it verifies that the convolutional layers focus on clinically relevant anatomical hallmark (such as osteophyte formation, joint space narrowing, and calcific deposits) rather than exploiting spurious background features or dataset artifacts; ii) it assists in diagnosing edge cases and failure modes where adjacent severity scores overlap. Crucially, in our multi-score setting, we extend this qualitative analysis to investigate the interplay and visual correlation between the two distinct pathological scales. By analysing the activation maps under varying severity levels of both KL and CPPD, we evaluate whether the independent backbones maintain high morphological fidelity to their respective target pathologies when they co-occur within the same knee joint region.

\subsubsection{KL \& CPPD relationship}\label{subsec:performance-relationship}

To assess whether the models successfully capture the asymmetric clinical relationship between KOA and CPPD, we implemented an evaluation framework based on Information Theory and paired statistical analysis. Rather than relying solely on classification accuracy, we evaluated the joint distribution of the predictions against the ground-truth clinical distribution.

Specifically, we measure the distribution fidelity using the Kullback-Leibler Divergence (KLD) \cite{kullback1997information} and an analysis of the residuals. Let $P$ be the true joint probability distribution (derived from the normalised counts of \Cref{fig:ground_truth}) and $Q$ be the predicted joint probability distribution. The KLD is formulated as:
\begin{equation}
D_\text{KL}(P \parallel Q)=\sum_{i,j} P(i,j) \log\left(\frac{P(i,j)}{Q(i,j)}\right)
\end{equation}
where $i$ and $j$ represent the KL and CPPD grades, respectively. Here, $P(i,j)$ represents the relative frequency (probability) of a patient exhibiting $\text{KL}=i$ and $\text{CPPD}=j$ according to the ground truth, with $Q(i,j)$ similarly defined for the model predictions. This metric quantifies the information lost when $Q$ is used to approximate $P$. A lower KLD indicates that the model predictions closely respect the clinical priors, heavily penalising biologically implausible predictions (e.g., combinations of grades that are non-existent in the clinical ground truth). 

To localise these discrepancies across the different severity levels, we also compute the residual matrix $R$ as the cell-wise difference between the true and predicted distributions, formulated as $R(i,j) = P(i,j) - Q(i,j)$. This residual matrix is linked to the KLD but with substantial differences: while $R(i,j)$ provides a linear, local measure of deviation (over- or under-estimation), the KLD serves as a global, non-linear aggregation of these residuals. In this way, due to the logarithmic ratio in the KLD formulation, the divergence heavily penalises strong negative residuals in regions where $P(i,j)$ is significant (i.e. when $Q(i,j) \to 0$ but the combination frequently occurs in clinical practice). Conversely, it suppresses biologically implausible predictions where $P(i,j) = 0$. This makes the KLD a much more sensitive and clinically relevant penalty than symmetric metrics as MAE.

Given that the models are evaluated across 20 independent random seeds, we constructed the predicted contingency table for each run and computed its KLD with respect to the true clinical distribution, together with the residuals with respect to the average contingency matrix. Finally, a paired non-parametric statistical pipeline (Kruskal-Wallis and Wilcoxon signed-rank test) is applied to assess whether the differences observed between the soft-labelling strategies and the nominal baseline are statistically significant.


\section{Results}\label{sec:results}

This section presents the results of the experiments described in \Cref{sec:proposed-methodology}, offering a comparison across soft-labelling techniques, analysing the results from a statistical perspective, and also examining the asymmetric correlation between CPPD and KL.

\subsection{Performance comparison across soft-labelling formulations}\label{subsec:softlabelling-results}
Table~\ref{tab:results} summarises the results obtained for CPPD and KL grading using the nominal formulation and the four considered soft-labelling strategies.

\begin{table*}[!htbp]
\centering
\tiny
\caption{Performance comparison ($Mean_{STD}$) across soft-labelling strategies on the KL and CPPD datasets. The best results are shown in bold, whereas the second-best results are shown in italics.}

\label{tab:results}
\begin{tabular}{llcccccc}
\toprule\toprule
Dataset & Soft-labelling & QWK & MAE & MS & BA & AMAE & MMAE \\
\midrule
CPPD & Nominal & $0.571_{0.177}$ & $0.726_{0.205}$ & $0.108_{0.086}$ & $0.446_{0.075}$ & $0.770_{0.219}$ & $1.262_{0.494}$ \\
CPPD & Binomial & $0.771_{0.026}$ & $0.488_{0.040}$ & $\mathit{0.406_{0.104}}$ & $0.563_{0.022}$ & $\mathit{0.470_{0.026}}$ & $\mathit{0.632_{0.115}}$ \\
CPPD & Beta & $\mathit{0.790_{0.030}}$ & $\mathit{0.453_{0.044}}$ & $\mathbf{0.454_{0.077}}$ & $\mathit{0.580_{0.031}}$ & $\mathbf{0.458_{0.032}}$ & $\mathbf{0.573_{0.084}}$ \\
CPPD & Triangular & $\mathbf{0.796_{0.018}}$ & $\mathbf{0.438_{0.025}}$ & $0.382_{0.063}$ & $\mathbf{0.585_{0.019}}$ & $0.471_{0.026}$ & $0.652_{0.064}$ \\
CPPD & Exponential & $0.757_{0.035}$ & $0.503_{0.048}$ & $0.405_{0.114}$ & $0.556_{0.030}$ & $0.478_{0.032}$ & $0.640_{0.113}$ \\
\midrule
CPPD & Average & $0.737_{0.057}$ & $0.522_{0.072}$ & $0.351_{0.089}$ & $0.546_{0.035}$ & $0.529_{0.067}$ & $0.752_{0.174}$ \\
\midrule
KL & Nominal & $0.616_{0.121}$ & $0.776_{0.207}$ & $0.156_{0.137}$ & $0.443_{0.064}$ & $0.784_{0.205}$ & $1.264_{0.455}$ \\
KL & Binomial & $0.744_{0.040}$ & $0.643_{0.062}$ & $0.138_{0.101}$ & $0.525_{0.041}$ & $0.585_{0.062}$ & $0.971_{0.165}$ \\
KL & Beta & $\mathbf{0.777_{0.022}}$ & $\mathbf{0.529_{0.031}}$ & $\mathbf{0.362_{0.085}}$ & $\mathbf{0.552_{0.023}}$ & $\mathbf{0.523_{0.031}}$ & $\mathbf{0.775_{0.136}}$ \\
KL & Triangular & $0.766_{0.031}$ & $0.548_{0.051}$ & $\mathit{0.346_{0.094}}$ & $0.542_{0.033}$ & $0.540_{0.048}$ & $0.792_{0.162}$ \\
KL & Exponential & $\mathit{0.773_{0.022}}$ & $\mathit{0.535_{0.040}}$ & $0.340_{0.097}$ & $\mathit{0.549_{0.038}}$ & $\mathit{0.528_{0.041}}$ & $\mathit{0.783_{0.117}}$ \\
\midrule
KL & Average & $0.735_{0.047}$ & $0.606_{0.078}$ & $0.268_{0.103}$ & $0.522_{0.040}$ & $0.592_{0.077}$ & $0.917_{0.207}$ \\
\bottomrule\bottomrule
\end{tabular}
\end{table*}

For CPPD, all soft-labelling approaches outperform the nominal baseline in terms of QWK, MAE, BA, AMAE, and MMAE. The highest QWK is obtained with the triangular distribution ($0.796_{0.018}$)), closely followed by the beta formulation ($0.790_{0.030}$). The lowest MAE is also achieved by the triangular approach ($0.438_{0.025}$), whereas the beta distribution attains the highest MS ($0.454_{0.077}$) and BA ($0.580_{0.031}$), together with the lowest AMAE ($0.458_{0.032}$) and MMAE ($0.573_{0.084}$). The exponential formulation yields intermediate performance, while the binomial distribution generally performs better than the nominal baseline but worse than the beta and triangular strategies.

For KL grading, a similar trend is observed. All soft-labelling formulations improve over the nominal setting. The best QWK is obtained with the beta distribution ($0.777_{0.022}$), followed by the exponential ($0.773_{0.022}$) and triangular ($0.766_{0.031}$) formulations. The beta approach also achieves the lowest MAE ($0.529_{0.031}$), the highest BA ($0.552_{0.023}$), and the lowest AMAE ($0.523_{0.031}$). The smallest MMAE is instead obtained with the triangular distribution ($0.792_{0.162}$), while the highest MS is provided by the beta formulation ($0.362_{0.085}$).

\begin{figure}[!htbp]
    \centering
    \includegraphics[width=\textwidth]{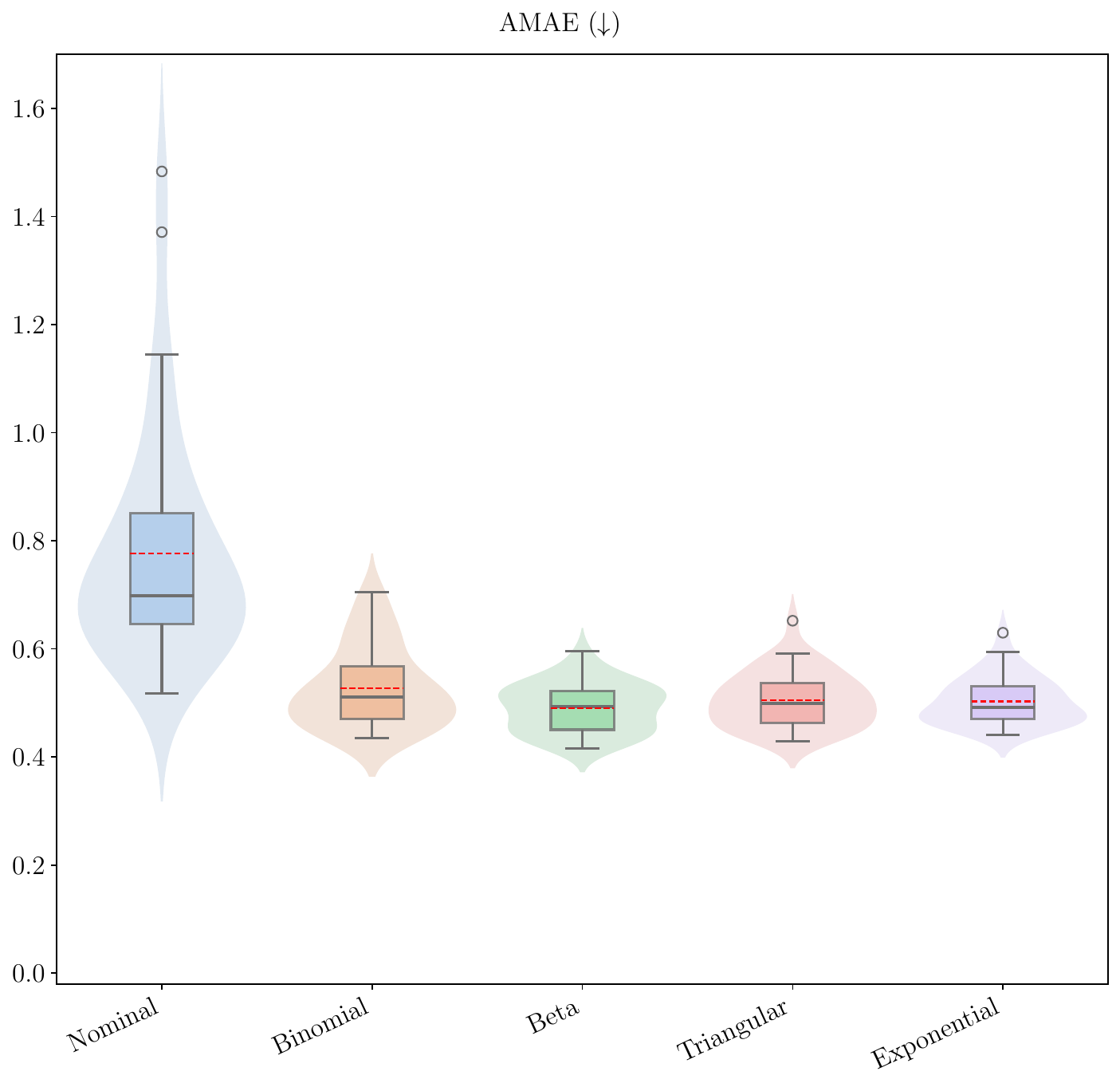}
    \caption{Combined violin and box plots of the AMAE distributions for all methodologies. The solid black line indicates the median, whereas the dashed red line represents the mean value for each methodology.}
    \label{fig:AMAE_boxplot}
\end{figure}

\Cref{fig:AMAE_boxplot} corroborates the quantitative results reported in Table~\ref{tab:results}. The beta formulation yields the lowest median AMAE together with the narrowest interquartile range, indicating both higher accuracy and greater robustness across the 20 repetitions. The triangular and exponential distributions also consistently outperform the nominal baseline, although with slightly larger variability and a greater number of outliers. In contrast, the nominal formulation exhibits both the highest AMAE values and the widest distribution, further confirming that one-hot encoding is less suitable for modelling the gradual transition between neighbouring KL and CPPD grades.

\begin{figure}[!htbp]
    \centering
    \includegraphics[width=\textwidth]{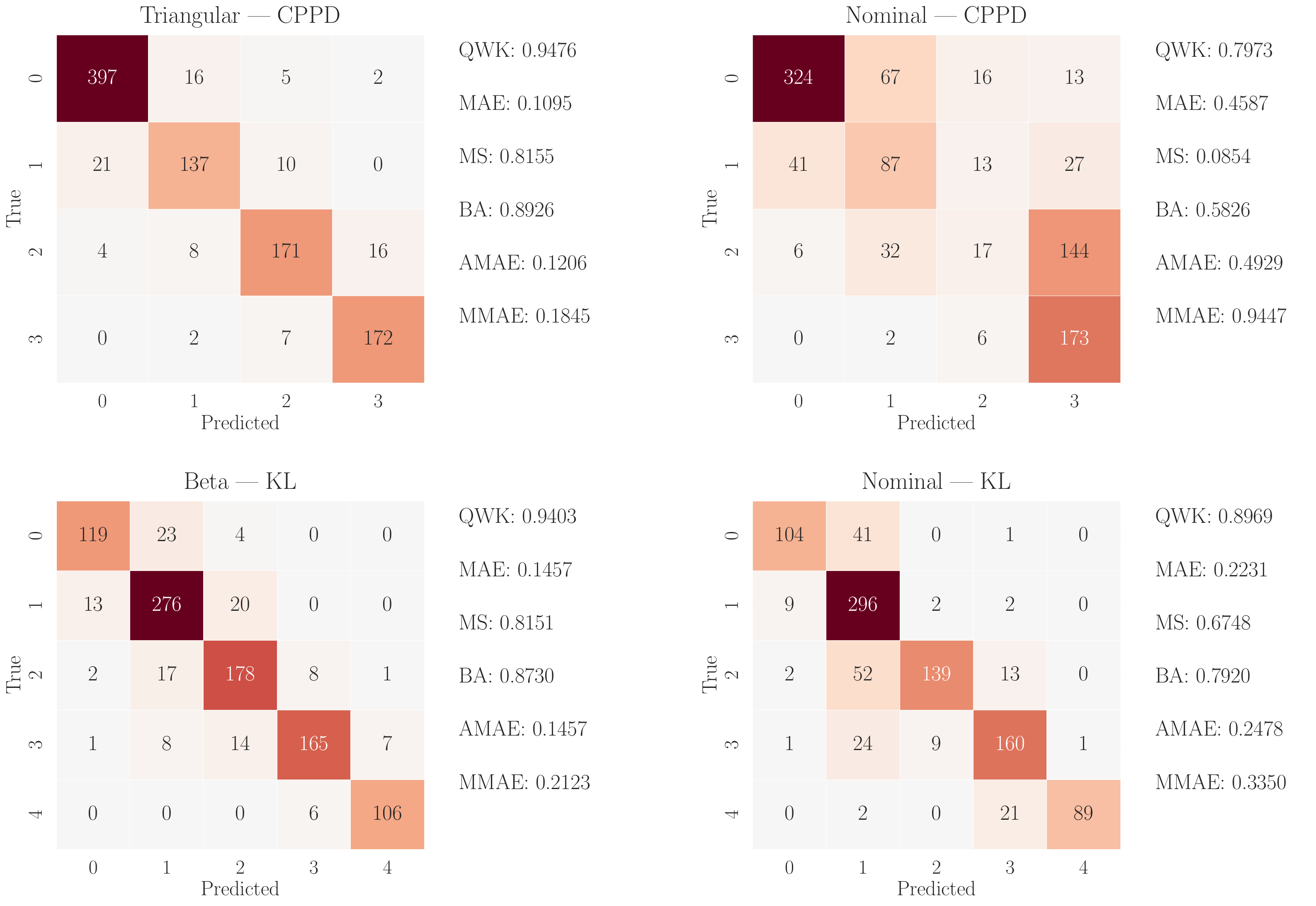}
    \caption{Confusion matrices comparing the best-performing soft-labelling models for each dataset with the nominal baseline for KL and CPPD grading.}
    \label{fig:confusion_matrices}
\end{figure}

\Cref{fig:confusion_matrices} reports the confusion matrices obtained by the best-performing soft-labelling model and by the nominal baseline for both KL and CPPD grading. This comparison provides additional insight into which severity grades benefit the most from the proposed formulation and whether the reduction in error is mainly due to fewer confusions between distant classes. Indeed, a closer inspection reveals that the soft-labelling approaches predominantly benefit the intermediate severity grades, which are inherently more susceptible to visual ambiguity. This is most pronounced in the CPPD dataset, where the nominal baseline almost entirely fails to recognise grade 2, misclassifying 144 of these instances as grade 3. The Triangular model successfully rectifies this behaviour, improving the true positive rate for grades 1 and 2.

A similar behaviour is observed in the KL dataset, where the Beta model reduces the nominal baseline's tendency to over-predict grade 1, leading to notable improvements for grade 2 and the extreme classes. Importantly, the confusion matrices show that the overall reduction in error is mainly driven by a decrease in misclassifications between distant classes. In both soft-labelling models, the remaining errors are mainly concentrated between adjacent classes. By limiting large grading discrepancies, the proposed formulation substantially reduces confusions spanning two or more grades, which directly contributes to the observed improvements in ordinal metrics such as the QWK and MAE.

\begin{figure}[!htpb]
    \centering
    \includegraphics[width=0.8\textwidth]{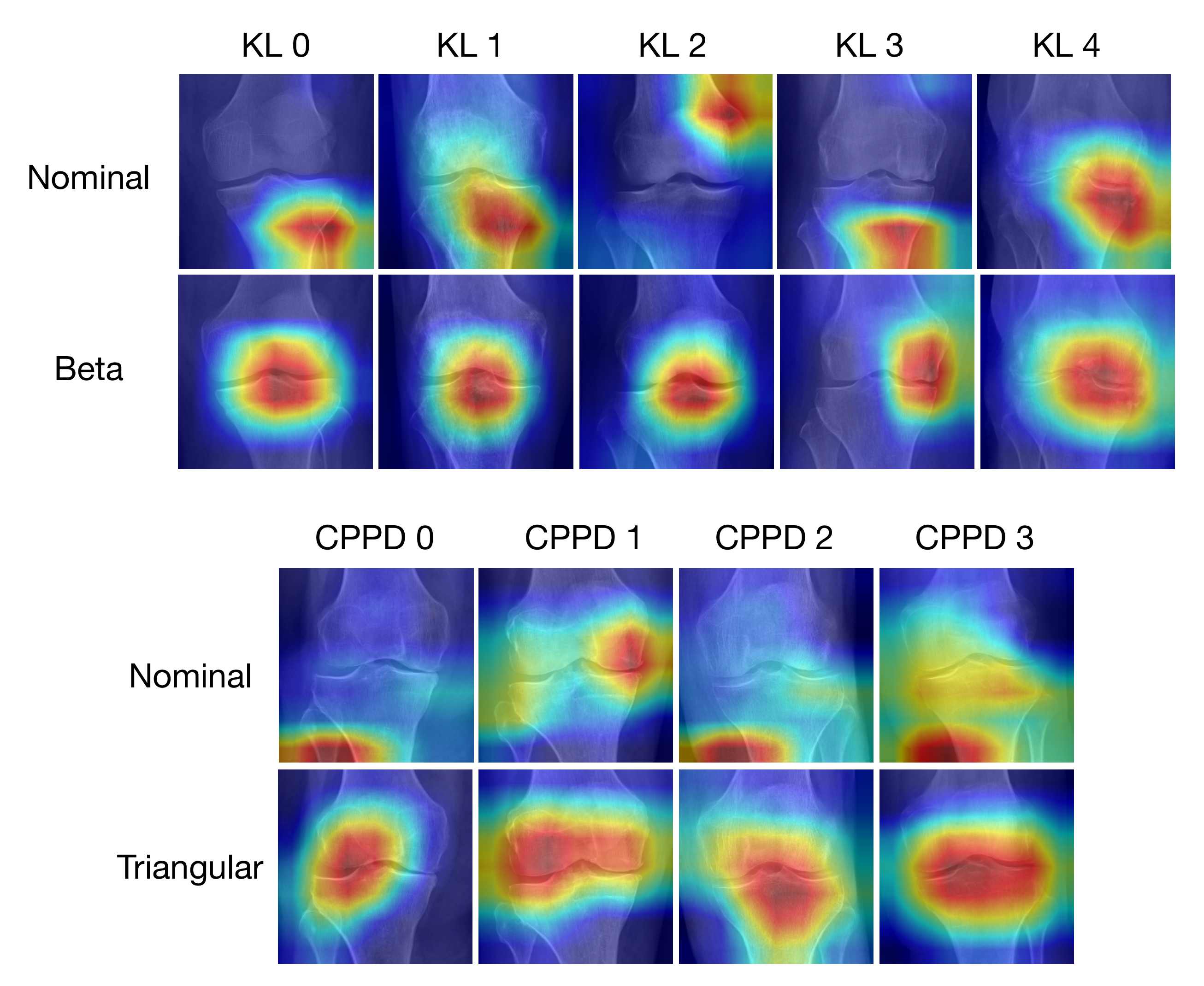}
    \caption{Grad-CAM visualisations across all severity grades for the nominal baseline and the best-performing soft-labelling models (Beta for KL and Triangular for CPPD).}
    \label{fig:gradcam}
\end{figure}

To visually inspect the localized effects of these performance improvements, \Cref{fig:gradcam} presents representative Grad-CAM visualizations across all severity grades, comparing the nominal baseline against our best-performing soft-labelling frameworks (Beta for KL and Triangular for CPPD). For KL grading, the Beta-supervised model cleanly tracks the progressive narrowing of the joint space and marginal osteophyte development across grades 0–4. Similarly, for CPPD grading, the Triangular-supervised model accurately isolates focal calcific deposits within the hyaline cartilage and fibrocartilage, scaling its attention area in strict accordance with the physical extension of the chondrocalcinosis.

\subsection{Statistical analysis}\label{subsec:statistical-analisis}

To assess whether the differences observed in AMAE were statistically significant, a two-way ANOVA was performed considering Models and Tasks as independent factors. The results, reported in \Cref{tab:anova_amae}, indicate a significant main effect of both Model ($p<0.001$) and Tasks ($p<0.001$) on the AMAE values. Conversely, the interaction between Model and Tasks was not significant (F$=1.289$), indicating that the relative behaviour of the different models remains consistent across the two grading tasks.

Following the ANOVA results, we applied Tukey's HSD post-hoc test to further explore pairwise differences between models (see \Cref{tab:tuckey_amae}). The analysis revealed two distinct subsets. The nominal approach forms a separate group with the worst AMAE value ($0.777$), indicating significantly worse performance compared to all soft-labelling methods. 

In contrast, all soft-labelling approaches (Binomial, Triangular, Exponential, and Beta) are grouped together, with no statistically significant differences among them. Within this group, the Beta distribution achieves the lowest AMAE value ($0.490$), followed closely by Exponential ($0.503$) and Triangular ($0.505$), while Binomial shows slightly higher error ($0.527$).

\begin{table}[!htpb]
    \centering
    \caption{Two-way ANOVA results for the AMAE metric. SS: sum of squares; DF: degrees of freedom; F: F-statistic.}
    \label{tab:anova_amae}
    \small
    \begin{tabular}{lcccc}
        \toprule\toprule
        \textbf{Factor} & \textbf{SS} & \textbf{DF} & \textbf{F} & \textbf{$p$-value} \\
        \midrule
        Model 
            & 2.367 
            & 4 
            & 57.961 
            & \(\mathbf{<0.001}\) \\
        Tasks 
            & 0.197 
            & 1 
            & 19.252 
            & \(\mathbf{<0.001}\) \\
        Model \(\times\) Tasks 
            & 0.053 
            & 4 
            & 1.289 
            & -- \\
        Residual 
            & 1.939 
            & 190 
            & -- 
            & -- \\
        \bottomrule\bottomrule
    \end{tabular}
\end{table}

\begin{table}[!htpb]
    \centering
    \footnotesize
    \setlength{\tabcolsep}{3pt}
    \renewcommand{\arraystretch}{1.0}
    \caption{Tukey test results for the AMAE metric.}
    \label{tab:tuckey_amae}
    \begin{tabular}{lcc}
        \toprule\toprule
         & \multicolumn{2}{c}{Subsets}\\\cmidrule{2-3}
        Group & S1 & S2\\
        \midrule
        Nominal     & $0.777$ &  \\
        Binomial    &  & $0.527$ \\
        Triangular  &  & $0.505$  \\
        Exponential &  & $0.503$ \\
        Beta        &  & $0.490$ \\
        \bottomrule\bottomrule
    \end{tabular}
\end{table}

\subsection{Asymmetric correlation between CPPD and KL}\label{sec:asymmetry}
In this section, we analyse the joint behaviour of the CPPD and KL grading systems in order to identify possible asymmetries between them. We first analyse the ground-truth contingency table of joint CPPD-KL annotations, and then examine how well the different prediction models reproduce this association. Finally, we use Kullback-Leibler divergence (KLD), together with formal statistical tests, to quantify differences between models and to assess whether the ordinal approaches provide a better description of the CPPD-KL relationship than the nominal one.

\begin{figure}[!htbp]
    \centering
    \includegraphics[width=\textwidth]{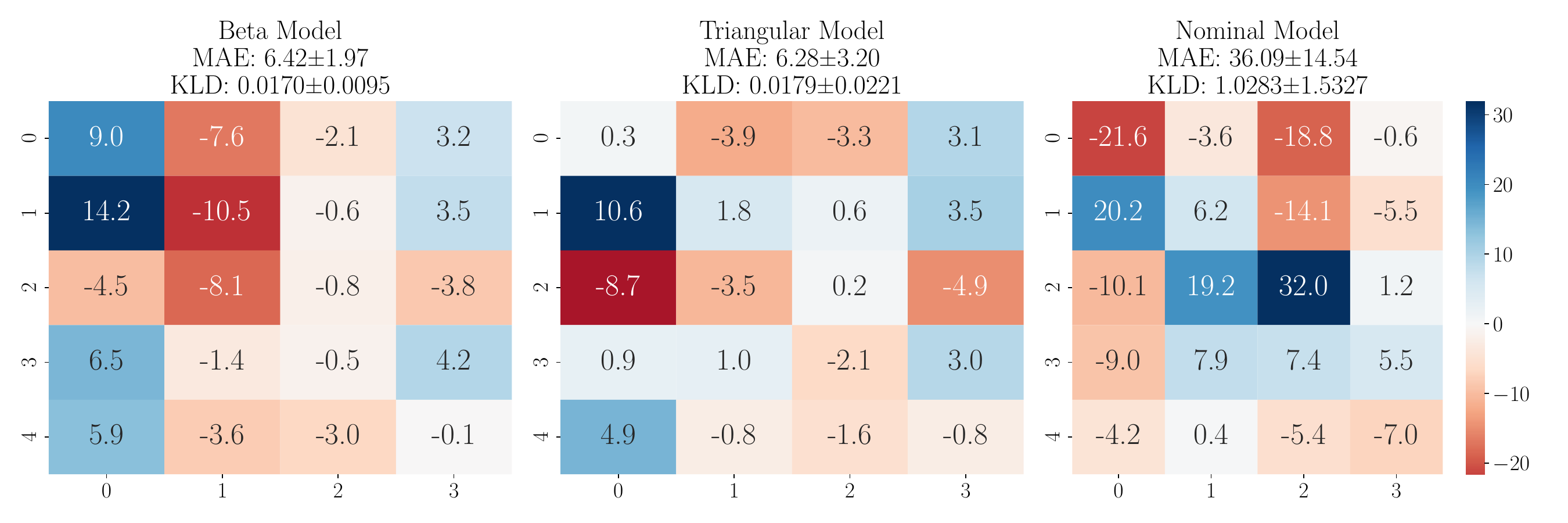}
    \caption{Residuals with the difference of the average obtained contingency tables and the ground-truth one (\Cref{fig:ground_truth}), by considering three models: nominal classifier and two ordinal ones (Beta and Triangular), together with average and standard deviation of $MAE$ and $KLD$}
    \label{fig:tests_kld}
\end{figure}

As discussed in \Cref{subsec:data-collection}, there is a clear asymmetric association between the KL and CPPD grading systems (see contingency table of \Cref{fig:ground_truth}). This inherent asymmetry serves as a benchmark to evaluate how each model captures the complex CPPD-KL relationship. In \Cref{fig:tests_kld}, each table depicts, for a given model, the residuals between the mean predicted contingency table across 20 runs and the true one included in \Cref{fig:ground_truth}. Positive residuals denote overestimation of the corresponding joint prediction made by the model, whereas negative residuals indicate underestimation. The resulting residual patterns, particularly in cells corresponding to mismatched CPPD and KL grades, show how each model deviates from the asymmetric association between the two scoring systems observed in the ground truth.

Above each residual heatmap, two summary measures are reported: MAE and KLD. KLD quantifies the divergence between the joint distribution given by the model predictions and that of the ground truth; larger KLD values correspond to greater discrepancies between the two distributions. The MAE, in contrast, acts as a global error on the contingency table counts: it is computed as the average absolute difference, cell by cell, between the predicted and true contingency tables.

From a qualitative point of view, the Triangular model attains the lowest MAE, indicating the smallest overall discrepancy in terms of absolute prediction errors. However, the Beta model exhibits the lowest KLD and therefore provides the closest approximation to the ground-truth joint distribution. The two ordinal models thus behave similarly, trading off slightly lower MAE (Triangular) against slightly lower divergence in KLD (Beta), whereas the nominal model shows clearly higher MAE and KLD values, reflecting a substantially poorer agreement with the ground truth.

\begin{table}[!htbp]
    \centering
    \caption{Statistical tests for KLD comparing ground truth contingency table from that obtained from the predictions of the nominal model and two ordinal ones (Beta and Triangular).}
    \label{tab:kld_stats}
    \begin{tabular}{@{}llc@{}}
        \toprule\toprule
        \textbf{Statistical test} & \textbf{Model(s)} & \textbf{$p$-value} \\
        \midrule
        
        \multirow{3}{*}{Normality (Shapiro)} 
        & Beta & $0.0023$ \\
        & Triangular & $< 0.0001$ \\
        & Nominal & $< 0.0001$ \\
        \midrule
        
        Global (Kruskal-Wallis) & All models & $1.59 \times 10^{-9}$ \\
        \midrule
        
        \multirow{3}{*}{Pairwise (Wilcoxon)} 
        & Beta vs. Nominal & $1.91 \times 10^{-6}$ \\
        & Triangular vs. Nominal & $1.91 \times 10^{-6}$ \\
        & Beta vs. Triangular & $0.1540$ \\
        
        \bottomrule\bottomrule
    \end{tabular}
\end{table}

To statistically support the comparison of models in terms of KLD, we performed a series of tests whose results are summarised in \Cref{tab:kld_stats}. First, Shapiro normality tests were applied separately to the KLD distributions of the Beta, Triangular, and Nominal models. The very small $p$-values ($p=0.0023$ for the Beta model and $p<0.0001$ for both the Triangular and Nominal models) indicate that none of them can be regarded as normal.

Consequently, we employed the Kruskal-Wallis test to assess global differences in KLD among the three models. The extremely small $p$-value ($p = 1.59 \times 10^{-9}$) indicates that the null hypothesis can be rejected, implying statistically significant overall differences in KLD across the models.

Given this global significance, we then carried out pairwise Wilcoxon signed-rank tests to compare models two by two. These tests reveal statistically significant differences in KLD between both soft-labelling models (Beta and Triangular) and the nominal model, whereas no significant difference is found between the Beta and Triangular models themselves.

\begin{figure}[!htpb]
    \centering
    \includegraphics[width=0.8\textwidth]{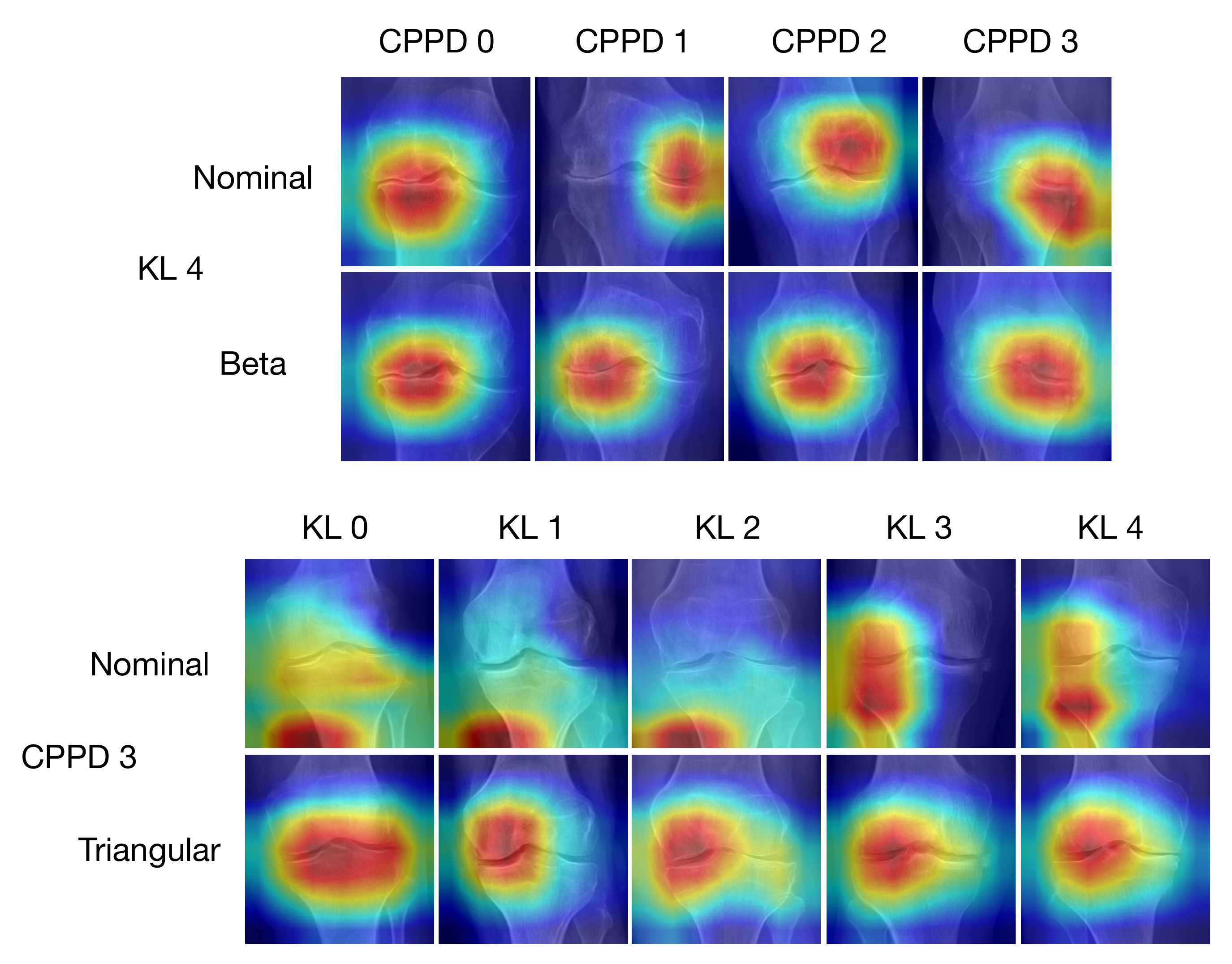}
    \caption{Grad-CAM analysis of model robustness under co-occurring pathologies. The evaluation isolates the highest severity level of one condition while varying the severity of the other.}
    \label{fig:gradcam_corr}
\end{figure}

We further assess the models' capability to handle co-occurring conditions using Grad-CAM. Specifically, we isolate the highest severity tier of each pathology, where structural changes are most dominant and clinically prominent, and systematically observe the behavior of the corresponding model's attention maps as the severity of the other condition varies.
\Cref{fig:gradcam_corr} illustrates this targeted evaluation. In the first scenario (top rows), we examine the KL-specific model on joints exhibiting maximum osteoarthritic degeneration (KL grade 4) across increasing levels of crystal deposition (CPPD grades 0 to 3). In the nominal baseline, the progressive introduction of chondrocalcinosis perturbs the feature extraction layers; as CPPD severity scales up, the nominal model's attention mass is progressively pulled away from the degenerative bone structures and misallocated toward focal calcifications. Conversely, the proposed Beta-supervised model successfully decouples the two signatures: it maintains a sharp, unwavering focus on the primary structural hallmarks of advanced osteoarthritis, such as severe joint space narrowing and subchondral sclerosis, remaining entirely unaffected by the increasing density of neighboring crystal matrices.
The second scenario (bottom rows) mirrors this analysis by fixing the CPPD model on maximum crystal extension (CPPD grade 3) while varying the background osteoarthritis framework from KL grade 0 to 4. The nominal baseline fails to isolate the CPPD deposits, expanding its receptive field inappropriately over the degenerative joint margins. In contrast, our Triangular-supervised soft-labelling formulation confines its activation maps strictly to the calcified hyaline cartilage zones, demonstrating great spatial and feature stability across the entire spectrum of co-occurring KL severity.

\section{Discussions}\label{sec:discussion}
This work addressed two clinically complementary yet pathophysiologically distinct grading tasks in knee radiographs: the KL scale, which reflects structural osteoarthritic degeneration, and the CPPD score, which captures the extent of calcium crystal deposition within cartilage and periarticular tissues. While KL grading is routinely used in clinical practice, CPPD assessment remains underreported and more challenging to standardise, particularly in the presence of advanced osteoarthritic changes where osteophytes, sclerosis, and joint space narrowing may obscure calcific deposits.

A central contribution of this work lies in modelling the intrinsic uncertainty of both grading systems through unimodal soft targets. From a clinical perspective, this is particularly relevant because the distinction between adjacent KL grades (e.g., grade 1 vs 2) or CPPD grades (e.g., sparse vs moderate calcifications) is often subtle and subject to inter-observer variability. By redistributing probability mass across neighbouring grades, soft-labelling better reflects the gradual progression of cartilage degeneration and crystal deposition, where sharp boundaries between stages rarely exist in real radiographic practice.
To systematically investigate how this uncertainty should be modelled, we evaluated four different unimodal distributions such as binomial, triangular, exponential, and beta, each encoding a distinct assumption on how diagnostic ambiguity spreads across adjacent grades. From a clinical standpoint, this is not a purely technical choice: different formulations reflect different hypotheses on disease progression. For instance, triangular distributions assume a symmetric and linearly decaying uncertainty around the annotated grade. In contrast, the beta distribution allows more flexible, potentially asymmetric uncertainty patterns, better capturing the heterogeneous and non-linear progressions. The binomial and exponential formulations instead provide more concentrated or distance-sensitive alternatives, enabling us to explore whether sharper or smoother transitions between grades better reflect the underlying radiographic variability.

The clinical relevance of this formulation is evident in the error patterns. Soft-labelling significantly reduces large-rank misclassifications (e.g., predicting KL 0 as KL 4 or CPPD 0 as CPPD 3), which are unlikely from a pathophysiological standpoint. Instead, errors are concentrated between adjacent grades, mirroring realistic diagnostic uncertainty. This is particularly important for intermediate stages, where radiographic signs such as early osteophytes or initial chondrocalcinosis may be faint or partially visible, and where even expert readers may disagree.

Importantly, the empirical results align with these modelling assumptions and reveal task-specific differences. The triangular distribution achieves the best performance for CPPD grading, suggesting that a linear uncertainty spread is well suited to modelling the progressive accumulation of calcific deposits, which tend to increase gradually in extent and visibility. In contrast, the beta distribution provides the most robust results for KL grading, likely due to its flexibility in capturing the heterogeneous progression of osteoarthritis, where structural changes (osteophytes, joint space narrowing, sclerosis, deformity) do not evolve uniformly across patients.

This task-specific modelling capacity is directly corroborated by our visual analysis (\Cref{fig:gradcam}). For both tasks, the nominal model frequently yields diffuse, scattered, or off-target attention maps, occasionally focusing on soft tissues or regions lacking diagnostic value. In contrast, the models trained under the proposed ordinal soft-labelling framework exhibit highly localised and anatomically consistent activation patterns. This sharper localisation demonstrates that regularising the target distribution with unimodal priors forces the latent feature representations to capture the true underlying morphology of disease progression, directly accounting for the substantial drop in large-rank misclassifications observed in the quantitative analysis.

From a statistical standpoint, ANOVA and post-hoc analyses confirm that these improvements are significant and attributable to the adoption of ordinal-aware supervision rather than random variability. Interestingly, no significant differences emerge among the different soft-labelling formulations, indicating that the primary advantage lies in the paradigm shift itself rather than in the specific choice of distribution. This suggests that even simple unimodal approximations are sufficient to capture the key properties of ordinal severity, without requiring complex modelling choices.

Beyond individual grading performance, this study provides insights into the clinical relationship between KL and CPPD. The observed asymmetric association (where low KL grades are strongly associated with absent CPPD, while high KL grades span the full spectrum of CPPD severity) highlights that osteoarthritic degeneration and crystal deposition represent partially independent processes. This is consistent with clinical observations: severe joint degeneration does not necessarily imply the presence of chondrocalcinosis, and conversely, CPPD can be present even in knees with mild or no radiographic osteoarthritis.

In this context, soft-labelling further demonstrates its value. Both beta and triangular models produce joint predictions that closely resemble the ground-truth distribution, as reflected by significantly lower KLD values. This suggests that ordinal supervision not only improves classification performance, but also enables the model to implicitly capture complex inter-score dependencies. In contrast, the nominal approach fails to reproduce this structure, leading to biologically less plausible predictions and higher divergence from the true distribution.

This is directly explained by the latent feature mechanics revealed in our dual-task visual stress-test (\Cref{fig:gradcam_corr}). By replacing rigid one-hot targets with unimodal probability distributions, soft-labelling strategy enables the separate backbones to implicitly capture complex, asymmetric inter-score dependencies without imposing an artificial, linear coupling between them.  As shown by the Grad-CAM maps, when the nominal model encounters highly confounded joint environments, its attention shifts erroneously between the two conditions, indicating that its latent space has entangled the distinct radiographic signatures. This feature entanglement directly accounts for the nominal model's poor distribution fidelity, leading to over-responsive, clinically less plausible joint predictions and a higher divergence from the true distribution.  On the contrary, the soft-labelled models treat co-occurring signs as distinct, independent visual layers. The KL-targeted network can process maximum structural narrowing without being distracted by adjacent crystal matrices, while the CPPD model successfully isolates extensive chondrocalcinosis even when surrounded by massive osteophytes and advanced bone deformities. This alignment between KLD and localised feature robustness highlighted by Grad-CAM confirms that our framework successfully replicates the diagnostic triage of experienced radiologists, ensuring that the automated multi-score assessment remains stable and dependable in complex, multi-pathology clinical scenarios. 

Despite these promising results, some limitations should be acknowledged. First, the dataset exhibits class imbalance, particularly for higher severity grades, which may still affect model performance despite the use of AMAE and balanced metrics. Second, KL and CPPD were modelled independently, and although their relationship was evaluated post hoc, a joint learning framework could further improve the modelling of their interaction. Finally, the study is based on retrospective data from a limited number of centres, and external validation on independent cohorts would be necessary to confirm generalisability.
Future work could explore multi-task learning strategies to jointly model KL and CPPD, as well as adaptive or data-driven soft-labelling schemes that dynamically adjust the uncertainty distribution based on sample difficulty. Additionally, extending the framework to other ordinal clinical scoring systems could further validate its general applicability.

\section{Conclusions}\label{sec:conclusions}

This study presented an ordinal deep learning framework for the automatic assessment of two clinically complementary radiographic scores, KL and CPPD, explicitly modelling their inherent uncertainty and graded progression. The results consistently demonstrate that replacing conventional one-hot supervision with soft ordinal targets leads to more accurate, robust, and clinically coherent predictions across both tasks.

Importantly, the benefit of this approach extends beyond improvements in standard performance metrics. By incorporating ordinal structure into the supervision signal, the proposed models reduce clinically implausible large-rank errors and instead favour predictions that reflect realistic diagnostic ambiguity, particularly in intermediate stages where radiographic interpretation is most challenging. This behaviour is crucial in the context of knee imaging, where subtle differences between adjacent grades often drive clinical decision-making. Furthermore, the ordinal formulation enables the model to better preserve the intrinsic relationship between osteoarthritic degeneration and crystal deposition. In particular, the proposed framework successfully captures the gradual and asymmetric association between KL and CPPD observed in the data, avoiding artificial coupling between the two scores and producing predictions that are more consistent with clinical reality.
From a methodological perspective, the proposed approach is simple, flexible, and easily integrable into existing deep learning pipelines, as it operates at the supervision level without requiring modifications to the underlying architecture. To the best of our knowledge, this is the first study to apply ordinal soft-labelling to both KL and CPPD grading from knee radiographs, highlighting the importance of explicitly modelling uncertainty in multi-score radiographic assessment.
Overall, this work supports the adoption of ordinal-aware learning strategies as a principled and clinically meaningful direction for improving automated severity assessment in musculoskeletal imaging, with potential applicability to a wide range of ordinal clinical scoring systems.

\section*{Acknowledgments}
The present study has been supported by the ``Agencia Estatal de Investigación (España)'' (grant ref.: PID2023-150663NB-C22 / AEI / 10.13039 / 501100011033), by the European Commission, AgriFoodTEF (grant ref.: DI\-GI\-TAL-2022-CLOUD-AI-02, 101100622), and by the Secretary of State for Digitalization and Artificial Intelligence ENIA International Chair (grant ref.: TSI-100921-2023-3). F. Bérchez-Moreno has been supported by ``Plan Propio de Investigación Submodalidad 2.2 Contratos predoctorales'' of the Universidad de Córdoba. Víctor Manuel Vargas has been supported by the Ministerio de Ciencia, Innovación y Universidades, the Agencia Estatal de Investigación and the European Social Fund Plus (grant ref.: MICIU / AEI / 10.13039 / 501100011033, JDC2024-054787-I).

\bibliographystyle{elsarticle-num}
\bibliography{bibliography}

\end{document}